\documentclass[10pt]{article}
\usepackage[margin=20mm]{geometry}
\usepackage[T1]{fontenc}
\usepackage{newtxtext,newtxmath}
\usepackage{graphicx,booktabs,microtype,xcolor}
\usepackage[numbers,sort&compress]{natbib}
\usepackage[hidelinks]{hyperref}
\usepackage{amsmath}
\newcommand{\ARXlinOneShort}{0.02}

\newcommand{\CalDith}{0.48}
\newcommand{\CalMaskH}{0.79}
\newcommand{\CalMaskQ}{0.57}
\newcommand{\CalNone}{0.65}
\newcommand{\CtwoEplateau}{0.57}
\newcommand{\CtwoSrange}{0.33$--$0.80}

\newcommand{\EBestOne}{0.01}

\newcommand{\ECtwoFbMax}{0.64}
\newcommand{\ECtwoFbMin}{0.10}

\newcommand{\ECtwoLag}{0.36}

\newcommand{\ECtwoLagOne}{0.44}

\newcommand{\ECtwoOne}{0.67}

\newcommand{\ECtwoWh}{0.80}

\newcommand{\EDithOne}{0.36}

\newcommand{\EDithWh}{0.49}

\newcommand{\EFTlinWh}{0.46}

\newcommand{\ENoiseArxHigh}{0.36}
\newcommand{\ENoiseArxZero}{0.00}
\newcommand{\ENoiseHigh}{0.37}
\newcommand{\ENoiseZero}{0.73}
\newcommand{\EOneEightK}{0.57}
\newcommand{\EOneOneTwoEight}{0.94}

\newcommand{\EPrbsOne}{0.50}
\newcommand{\ESineOne}{0.95}
\newcommand{\ESineOneArx}{0.84}
\newcommand{\ETfmFbMax}{0.72}
\newcommand{\ETfmFbMin}{0.02}

\newcommand{\ETfmLagMax}{0.87}
\newcommand{\ETfmLagMin}{0.40}

\newcommand{\ExcRhoMax}{0.96}
\newcommand{\ExcRhoMin}{0.64}

\newcommand{\FTminutes}{26}

\newcommand{\KDithGain}{0.84}
\newcommand{\KDithRho}{0.85}

\newcommand{\KGainOne}{0.41}

\newcommand{\KNoneGain}{0.41}
\newcommand{\KNoneRho}{0.69}

\newcommand{\KRhoOne}{0.69}

\newcommand{\KoneShare}{90}

\newcommand{\NArxSb}{0.077}

\newcommand{\NCtwoSb}{0.544}
\newcommand{\NCtwoWh}{0.556}

\newcommand{\NFTCt}{0.428}
\newcommand{\NFTSb}{0.629}
\newcommand{\NFTWh}{0.182}

\newcommand{\NFTwSb}{0.443}

\newcommand{\NTfmLagWh}{0.201}

\newcommand{\NTfmWh}{1.081}
\newcommand{\OracleGainOne}{0.56}
\newcommand{\RhoFactWhatif}{0.92}

\newcommand{\SCtwoOne}{0.50}

\newcommand{\SDithOne}{0.84}

\newcommand{\SFTlinWh}{0.92}

\newcommand{\STfmLagMax}{1.08}
\newcommand{\STfmLagMin}{0.81}

\newcommand{\TFMmemR}{1.000}
\newcommand{\TabMaxR}{0.93}
\newcommand{\TabMinR}{0.71}
\newcommand{\TruthMaxR}{0.04}

\newcommand{\BADuf}{0.14}

\newcommand{\CtwoStwoMax}{0.70}
\newcommand{\CtwoStwoMin}{0.20}
\newcommand{\CtwoSzeroMax}{0.87}
\newcommand{\CtwoSzeroMin}{0.48}
\newcommand{\ECtwoBwOnly}{0.50}
\newcommand{\EDithBwOnly}{0.34}

\newcommand{\FTSrangeM}{0.83$--$0.96}
\newcommand{\FTaltDufTwo}{0.70}
\newcommand{\FTbeatsICLours}{3}

\newcommand{\FTmDuf}{0.38}

\newcommand{\FTmOne}{0.17}

\newcommand{\FTmWh}{0.20}

\newcommand{\HxNarrowDufCtwo}{0.71}
\newcommand{\HxNarrowDufFT}{0.42}

\newcommand{\HxNarrowLagCtwo}{0.63}
\newcommand{\HxNarrowLagFT}{0.17}
\newcommand{\HxNarrowOneArx}{0.10}
\newcommand{\HxNarrowOneCtwo}{0.79}
\newcommand{\HxNarrowOneFT}{0.34}

\newcommand{\HxSineLagArx}{0.78}
\newcommand{\HxSineLagCtwo}{0.96}
\newcommand{\HxSineLagFT}{0.23}

\newcommand{\IoursWh}{0.22}
\newcommand{\MBCt}{OE}

\newcommand{\MBWh}{NARX}
\newcommand{\MBvCt}{0.495}

\newcommand{\MBvWh}{0.125}
\newcommand{\MDdCt}{\ensuremath{-}0.328}
\newcommand{\MDdSb}{\ensuremath{-}0.042}
\newcommand{\MDdWh}{\ensuremath{-}0.261}
\newcommand{\MDhiCt}{\ensuremath{-}0.128}
\newcommand{\MDhiSb}{\ensuremath{-}0.020}
\newcommand{\MDhiWh}{\ensuremath{-}0.167}
\newcommand{\MDloCt}{\ensuremath{-}0.547}
\newcommand{\MDloSb}{\ensuremath{-}0.049}
\newcommand{\MDloWh}{\ensuremath{-}0.310}
\newcommand{\MFTdCt}{+0.021}

\newcommand{\MFTdWh}{+0.064}
\newcommand{\MFThiCt}{+0.207}

\newcommand{\MFThiWh}{+0.084}
\newcommand{\MFTloCt}{\ensuremath{-}0.225}

\newcommand{\MFTloWh}{+0.049}
\newcommand{\MaseBase}{0.69}
\newcommand{\MaseFT}{0.89}
\newcommand{\MaseFTw}{0.82}

\newcommand{\MaseWorsePct}{82}

\newcommand{\NLbetter}{5}
\newcommand{\NLbetterB}{4}
\newcommand{\NLcells}{8}
\newcommand{\NLtie}{2}
\newcommand{\NLworse}{1}
\newcommand{\NSeeds}{3}
\newcommand{\NftdithE}{0}
\newcommand{\NftdithWorse}{4}
\newcommand{\NsinhE}{2}
\newcommand{\NsinhS}{5}

\newcommand{\NubandWorse}{6}
\newcommand{\NuwhiteE}{4}
\newcommand{\ParrotCopy}{0.16}
\newcommand{\ParrotTruth}{0.78}
\newcommand{\QmeanMaxdS}{0.03}
\newcommand{\ReplayText}{Fine-tuning with an equal number of univariate M4 weekly series mixed into the corpus (FTmix; a different frequency from the hourly test data) gives a median MASE of 0.97 on the hourly series (base 0.69, FT 0.89). Its median what-if error at $L=1024$ is within 0.05 of FT on 5 of 6 classes (lag+delay 0.10, 1-DOF 0.19, 2-DOF 0.30, Duffing 0.42, Wiener--Hammerstein 0.25, Coulomb 0.31). Replay therefore largely kept the what-if gains but did not prevent forgetting; it made univariate forecasting worse still (one run). Preventing forgetting remains open.}
\newcommand{\RiArxm}{0.074}
\newcommand{\RiCtwom}{0.766}
\newcommand{\RiCtwos}{0.381}
\newcommand{\RiFTm}{0.268}
\newcommand{\RiFTs}{0.493}
\newcommand{\RiFTwm}{0.329}
\newcommand{\RkMax}{0.96}
\newcommand{\RkMin}{0.35}
\newcommand{\SeedRangeMax}{0.04}
\newcommand{\SenSCou}{0.48}
\newcommand{\SenSDuf}{0.74}
\newcommand{\SenSWh}{0.73}
\newcommand{\SevArx}{0.051}

\newcommand{\SevCtwo}{0.303}

\newcommand{\SevFT}{0.195}

\newcommand{\SevFTd}{+0.108}
\newcommand{\SevFThi}{+0.153}
\newcommand{\SevFTlo}{+0.055}

\newcommand{\SevGCtwo}{+0.04}

\newcommand{\SevGFT}{+0.41}

\newcommand{\SevOe}{0.045}

\newcommand{\SevTfm}{0.312}

\newcommand{\SsinhOne}{0.76}

\newcommand{\Eint}{E_{\mathrm{int}}}

\title{Attenuated in-context identification in time-series foundation models: diagnosis under counterfactual inputs and repair by synthetic forced-system fine-tuning}
\author{Hong-In Won\thanks{Corresponding author: \texttt{luvhayym@kitech.re.kr}. ORCID 0000-0003-1609-8447.}\\[4pt]
\small Manufacturing AI Research Center, Korea Institute of Industrial Technology, Incheon, Republic of Korea\\
\small HYU-KITECH Joint Department, Graduate School, Hanyang University, Seoul, Republic of Korea}
\date{}

\begin{document}
\maketitle

\begin{abstract}
Covariate-aware time-series foundation models (TSFMs) promise training-free what-if answers for instrumented plants: the change in output that a different future input would cause. We test this on forced engineering systems with exact counterfactuals, comparing Chronos-2, TimesFM-2.5 and TabPFN-TS with classical system identification fitted to the same context. Through their default covariate interfaces, TimesFM-2.5 and TabPFN-TS are memoryless: the predicted effect of an input change is a same-time function of that change ($R^2 = \TFMmemR$ for TimesFM-2.5). Chronos-2 identifies dynamics in context but attenuates them. Its predicted effect is $\CtwoSrange$ of the true effect, its recovered impulse response has the wrong shape, and its error on a one-degree-of-freedom oscillator levels off at $\CtwoEplateau$ with 8192 context samples, where ARX fitted to 256 samples reaches $\ARXlinOneShort$. Context dither at inference lowers the what-if error on all six synthetic classes without training. A \FTminutes-minute fine-tune on synthetic forced systems restores the response magnitude (sensitivity $\FTSrangeM$) and outperforms structure-agnostic identification on Wiener--Hammerstein and a held-out friction class. A specialised in-context identifier trained on the same data comes close, so the forced-system data carry most of the gain. On three of four measured plants classical identification remains clearly better, and the fine-tuned model loses part of its univariate forecasting skill. Paired counterfactual inputs, together with shuffled future inputs on measured records, test two properties: whether the covariate interface can represent dynamics and whether the pretraining prior covers the plant's time scale. Only the counterfactual pairs expose the attenuation.
\end{abstract}

\section{Introduction}

A model that forecasts a plant well is not necessarily a model that can answer what-if questions about it. Model-based control, scheduling and design exploration all require the change in output caused by a change of the input, not only the most likely continuation of the recorded trajectory. Time-series foundation models (TSFMs) now accept known future covariates and are evaluated on benchmarks that include them~\citep{ansari2025chronos2,das2024timesfm,hoo2025tabpfnts}. If such a model has learned to identify the input--output dynamics of a system from its context window, it would provide a training-free what-if model for any instrumented plant. Whether it does has not been tested on systems where the counterfactual answer is known.

Recent work points in both directions. Chronos-2 learns cross-variate structure in context through group attention and leads covariate forecasting benchmarks~\citep{ansari2025chronos2}. Controlled studies show that Chronos-2 and TabPFN-TS capture simple static target--covariate relations~\citep{berthelier2026covariates}, that TSFMs over-weight persistence when the generator of a series is intervened on~\citep{jander2026causal}, and that on chaotic systems they may copy context segments~\citep{zhang2024zeroshot,zhang2025parroting}. For building heating, ventilation and air-conditioning (HVAC) control, a covariate TSFM gave poor intervention effects, and a separately identified forced-response operator was needed~\citep{wang2026thermoforce}. Specialised transformers trained on synthetic linear and Wiener--Hammerstein systems can simulate a system from an input--output context~\citep{forgione2023incontext,rufolo2024probabilistic,piga2024adapting}. None of these studies measures the counterfactual response of general-purpose TSFMs on forced engineering systems against classical identification fitted to the same data, nor asks which component of the model limits it.

We make this measurement. Our test systems are synthetic plants with exact counterfactuals and four measured plants: three public nonlinear benchmarks and a base-excitation rig of our own. The synthetic plants are a first-order lag with dead time; oscillators with one and two degrees of freedom (DOF), denoted 1-DOF and 2-DOF; Duffing, Wiener--Hammerstein and Coulomb-friction systems; and a Bouc--Wen hysteresis class used only for testing. For each context we change the future input by scaling, removing, stepping, replacing it with a resonant sine or with a fresh realisation, and we score the predicted change of output against the true change. The comparison baselines are autoregressive models with exogenous input (ARX), output-error, subspace (N4SID) and polynomial, simulation-error and neural nonlinear ARX (NARX) models identified from the same context~\citep{ljung1999system,vanoverschee1994n4sid,billings2013narmax}, a nearest-neighbour context-copy model, a specialised in-context identification transformer, and a known-structure grey-box model as a bound.

Our contributions are:
\begin{enumerate}
\item A diagnosis of three covariate-aware TSFMs under what-if inputs. Two have memoryless default covariate interfaces and, through them, cannot represent a dynamic effect. The third, Chronos-2, identifies the dynamics but attenuates and distorts the response. The attenuation saturates with context length, follows the classical excitation requirements, and decreases when the context output is noisier (Section~\ref{sec:diagnosis}).
\item A training-free remedy: context dither at inference, which lowers what-if error on all synthetic classes and factual error on three public measured benchmarks (Section~\ref{sec:dither}).
\item Evidence that the attenuation is mainly a deficit of the pretraining prior rather than of model capacity. A short fine-tune on synthetic forced systems largely removes it and makes the model competitive with structure-agnostic identification on nonlinear plants. A specialised identifier trained on the same data comes close, so the data rather than the backbone carry most of the gain. On most measured plants the adapted model still trails classical identification, partly because a plant's time scale can fall outside the synthetic prior, and it forgets part of its univariate forecasting skill (Section~\ref{sec:adapt}).
\end{enumerate}

\section{Problem and protocol}\label{sec:protocol}

\begin{figure}[t]
\centering
\includegraphics[width=\textwidth]{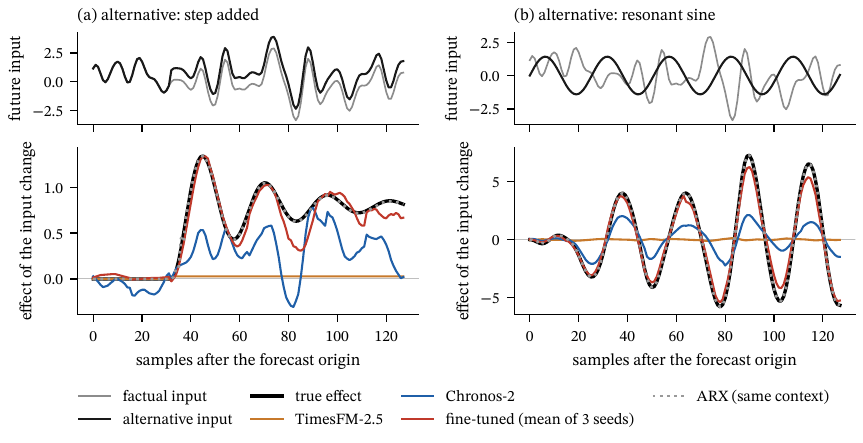}
\caption{One what-if query on a 1-DOF oscillator ($L = 1024$; median Chronos-2 error of 30 instances). Top: factual future input $u$ and alternative $u'$ (a: step added at $t = H/4$; b: resonant sine). Bottom: true effect $\Delta y$ and predicted effects $\Delta\hat y$. Fine-tuned: Chronos-2 after the fine-tune of Section~\ref{sec:adapt}, mean of three seeds.}
\label{fig:whatif}
\end{figure}

\paragraph{Paired counterfactual queries.} A context of length $L$ holds the input $u_{1:L}$ and output $y_{1:L}$ of a plant, and the future input $u_{L+1:L+H}$ is known. A model $m$ returns a forecast $\hat y(u_{L+1:L+H})$. For an alternative future input $u'$ the true effect is $\Delta y = y(u') - y(u)$, computed from the same plant state at time $L$, and the predicted effect is $\Delta\hat y = \hat y(u') - \hat y(u)$. The free response and any forecast bias shared by the two calls cancel. We report the what-if error $\Eint = \|\Delta\hat y - \Delta y\|/\|\Delta y\|$ (0 is exact; 1 equals predicting no change) and the sensitivity ratio $S = \|\Delta\hat y\|/\|\Delta y\|$. Six alternatives are scored per context (input scaled by 0.5 and 2, zero input, a step added, a resonant sine, a fresh input realisation). The instance score is the median over the six, and the system score is the median over 30 instances with a bootstrap interval. The horizon is $H = 128$ samples. Fig.~\ref{fig:whatif} shows one such query. In this instance ARX fitted to the same context nearly coincides with the true effect, TimesFM-2.5 predicts almost no change, and Chronos-2 has the right direction but too small a magnitude.

\paragraph{Systems.} Six synthetic classes, and a Bouc--Wen hysteresis class~\citep{wen1976hysteretic} used only for testing, are simulated with a fourth-order Runge--Kutta method under zero-order-hold input, with parameters sampled per instance (Appendix~\ref{app:systems}). The context input is Gaussian noise band-limited to three times the dominant frequency (cutoff clipped to 0.08--0.4 cycles per sample; Appendix~\ref{app:systems}). Context outputs are standardised and carry 3\% white noise; the future truth is noise-free. Four measured benchmarks come from the nonlinear system identification collection~\citep{schoukens2026benchmarking,wigren2013three,schoukens2017three}: Silverbox, Wiener--Hammerstein, Cascaded Tanks and the electro-mechanical positioning system (EMPS). They have no counterfactual truth, so we score the factual forecast by its normalised root-mean-square error (NRMSE) and the \emph{shuffle gain}: the increase of the error when the true future input is replaced by the input of another window. A model that ignores the input has zero shuffle gain. EMPS is excluded: the record is a periodic closed-loop tracking experiment whose output repeats every 624 samples (a context-copy model reaches zero error on it), so it cannot test input response. We also use a base-excitation rig of our own: a beam in stick-slip contact driven by a 10--100\,Hz sweep, with base acceleration as input and beam velocity as output (the source data of a separate study, used here for evaluation only).

\paragraph{Models.} Chronos-2 receives the input as a past-and-future covariate. TimesFM-2.5 (200M) uses its in-context regression (XReg) followed by TimesFM on the residual, with one regression per context. TabPFN-TS uses the publicly released TabPFN~v2 regression weights~\citep{hollmann2025tabpfn}; newer licence-gated weights were not evaluated. All models run locally, and all output is the median forecast.

\paragraph{Baselines on the same context.} ARX, output-error (OE) and polynomial NARX models (degree $\le 3$) are fitted with orders selected by simulation error on the last quarter of the context and then refitted on the whole context. N4SID is fitted with its order chosen the same way. For each system and context length we report the best of these on the evaluation instances, an oracle selection that favours the baseline. We also include a nearest-neighbour context-copy model~\citep{zhang2025parroting} and the in-context identification transformer of Forgione et al.~\citep{forgione2023incontext} (ICL-TF), using its linear and Wiener--Hammerstein checkpoints; it reads the most recent $\min(L, 400)$ context samples, its training context length. For the nonlinear classes we add a NARX model refined on free-run simulation error (SE), denoted SE-NARX, a NARX model built on a multilayer perceptron (MLP), denoted MLP-NARX, and a grey-box model with the true equations of motion and unknown parameters; the grey-box model is a known-structure bound rather than a competitor. To separate the effect of training data from that of the backbone, we also train the in-context identification transformer on the same synthetic series used to fine-tune Chronos-2 (Section~\ref{sec:adapt}).

\section{Diagnosis of the covariate interfaces}\label{sec:diagnosis}

\begin{figure}[t]
\centering
\includegraphics[width=0.8\textwidth]{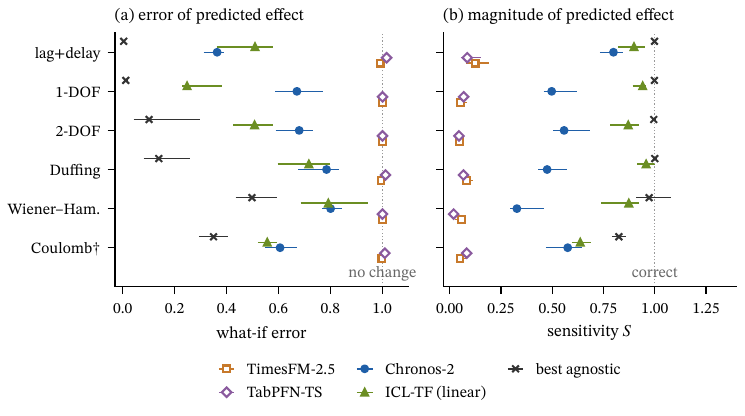}
\caption{What-if error and sensitivity of three TSFMs, the published in-context identification transformer (linear checkpoint) and the best structure-agnostic identifier (best of ARX, OE, N4SID, polynomial NARX, SE-NARX and MLP-NARX) on six synthetic classes ($L=1024$, 30 instances). Medians with 95\% bootstrap intervals. $^\dagger$Held out from all fine-tuning.}
\label{fig:overview}
\end{figure}

\begin{table}[t]
\centering\small
\caption{Median what-if error $\Eint$ at $L=1024$ (30 instances, six alternative inputs each). ICL-TF: published in-context identification transformer~\citep{forgione2023incontext}, better of its linear and Wiener--Hammerstein checkpoints; ICL-TF$^\ast$: the same architecture trained on the FT training series. +dither: Chronos-2 with context dither (Section~\ref{sec:dither}). FT: Chronos-2 fine-tuned on synthetic forced systems (median over instances of the per-instance mean over seeds, $\pm$ half-range of the seed medians); FT-wide: FT with a time-scale-randomised prior (Section~\ref{sec:adapt}). ICL-TF and ICL-TF$^\ast$ read the last $\min(L, 400)$ context samples. Best agnostic: best of ARX, OE, N4SID, polynomial NARX, SE-NARX and MLP-NARX on the evaluation instances. Known struct.: grey-box model with the true equations. $^\dagger$Held out from fine-tuning. Table~\ref{tab:main256} gives $L=256$.}
\label{tab:main}
\resizebox{\textwidth}{!}{\begin{tabular}{lcccccccccc}
\toprule
system & TimesFM-2.5 & TabPFN-TS & Chronos-2 & +dither & ICL-TF & ICL-TF$^\ast$ & FT & FT-wide & best agnostic & known struct. \\
\midrule
lag+delay & 0.99 & 1.02 & 0.36 & 0.29 & 0.51 & 0.09 & 0.10$\pm$0.01 & 0.09 & 0.00 (NARX) & -- \\
1-DOF & 1.00 & 1.00 & 0.67 & 0.36 & 0.25 & 0.26 & 0.17$\pm$0.01 & 0.19 & 0.01 (ARX) & -- \\
2-DOF & 1.00 & 1.00 & 0.68 & 0.55 & 0.51 & 0.39 & 0.24$\pm$0.01 & 0.27 & 0.10 (N4SID) & -- \\
Duffing & 0.99 & 1.01 & 0.79 & 0.59 & 0.70 & 0.40 & 0.38$\pm$0.02 & 0.43 & 0.14 (SE-NARX) & 0.00 \\
Wiener--Ham. & 1.00 & 1.00 & 0.80 & 0.49 & 0.69 & 0.22 & 0.20$\pm$0.01 & 0.24 & 0.50 (SE-NARX) & 0.00 \\
Coulomb$^\dagger$ & 1.00 & 1.01 & 0.61 & 0.45 & 0.56 & 0.36 & 0.30$\pm$0.01 & 0.34 & 0.35 (OE) & 0.00 \\
Bouc--Wen$^\dagger$ & -- & -- & 0.50 & 0.34 & -- & -- & 0.23$\pm$0.01 & 0.24 & 0.28 (NARX) & -- \\
\bottomrule
\end{tabular}}
\end{table}

\paragraph{Memoryless default covariate interfaces.} TimesFM-2.5 and TabPFN-TS leave the forecast almost unchanged when the future input changes. Their sensitivity is below 0.15 on every system, and their what-if error is 1.00 to within 0.02 (Fig.~\ref{fig:overview}, Table~\ref{tab:main}). The cause is the default interface, not a small effect size. TimesFM-2.5 reads covariates through an in-context linear regression on same-time values, so its predicted effect is an exact same-time linear function of the input change ($R^2 = \TFMmemR$ on every system, by construction). The true effect is not ($R^2 \le \TruthMaxR$), because it is carried by the plant's memory. TabPFN-TS with its default features behaves similarly ($R^2$ between \TabMinR{} and \TabMaxR). On the measured benchmarks both models have near-zero shuffle gain ($-0.01$ to $+0.01$; Table~\ref{tab:bench}). Giving the interface memory confirms that the interface is the limit rather than the backbone. With 32 lagged copies of the input as additional covariates, TimesFM-2.5 stops ignoring the input: its sensitivity rises to \STfmLagMin--\STfmLagMax, and its error on the measured Wiener--Hammerstein benchmark falls from \NTfmWh{} to \NTfmLagWh. TabPFN-TS with the same lags is no longer memoryless either (same-time $R^2 \le 0.05$). Neither becomes competitive on the synthetic classes (TimesFM-2.5: \ETfmLagMin--\ETfmLagMax), because an unregularised regression on many correlated columns is a poor impulse-response identifier (Appendix~\ref{app:e4}).

\paragraph{Attenuated identification in Chronos-2.} Chronos-2 is the only TSFM whose default forecast depends on the input history. Its predicted effect has the right direction but only $\CtwoSrange$ of the true magnitude, and its what-if error ranges from \ECtwoLag{} (lag+delay) to \ECtwoWh{} (Wiener--Hammerstein). It does not copy the context: at $L=1024$ its factual forecast correlates with the truth at \ParrotTruth{} on average and with a nearest-neighbour continuation at \ParrotCopy. It loses to classical identification fitted to the same context on every system at both context lengths. The loss is largest on 1-DOF (\ECtwoOne{} against \EBestOne), where ARX is the correct model class, and is also large on 2-DOF and Duffing (Table~\ref{tab:main}). The attenuation is present for every alternative input, including the zero input (sensitivity \CtwoSzeroMin--\CtwoSzeroMax), and strongest for the doubled input on every class except lag+delay, where the step is lower (sensitivity \CtwoStwoMin--\CtwoStwoMax; Table~\ref{tab:alt}). Three input-handling explanations were checked. The model standardises covariates with context statistics only, so changing the future input cannot rescale its history. The mean of the predicted quantiles gives the same sensitivity as the median (largest difference \QmeanMaxdS), so median shrinkage does not explain it. Chronos-2 also applies an $\operatorname{arcsinh}$ compression after standardisation, which shrinks large input excursions. Feeding $\sinh(u)$ instead of $u$, which approximately undoes this compression, raises the sensitivity on \NsinhS{} of 6 classes (1-DOF: \SCtwoOne{} to \SsinhOne) but lowers the error on only \NsinhE. The compression therefore explains part of the magnitude deficit and none of the shape error.

\begin{figure}[t]
\centering
\includegraphics[width=\textwidth]{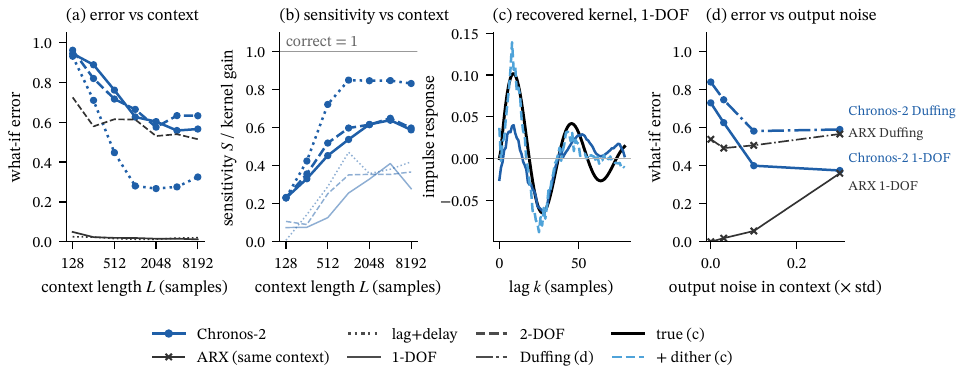}
\caption{Chronos-2 in-context identification (medians over a separate set of 30 instances per class, five alternatives including a pulse; Appendix~\ref{app:systems}). (a) What-if error versus context length, with ARX on the same contexts. (b) Sensitivity (thick) and impulse-response gain (thin). (c) Impulse response from a what-if pulse, 1-DOF, $L=4096$, median-error instance. (d) What-if error versus context output noise ($L=1024$).}
\label{fig:mechanism}
\end{figure}

\paragraph{Context length and response shape.} The studies of Fig.~\ref{fig:mechanism} use a separate set of 30 instances per class with five alternative inputs, one of them a pulse (Appendix~\ref{app:systems}), so their values differ slightly from Table~\ref{tab:main}. More context helps only up to about 2000 samples (Fig.~\ref{fig:mechanism}a). On the 1-DOF oscillator the error falls from \EOneOneTwoEight{} at 128 samples to \EOneEightK{} at 8192, where ARX fitted to 256 samples of the same contexts reaches \ARXlinOneShort. For lag+delay the error even rises again beyond 4096 samples. At long contexts the sensitivity levels off near 0.6 on the oscillators and 0.85 on lag+delay instead of approaching 1 (Fig.~\ref{fig:mechanism}b). A what-if pulse exposes the impulse response that Chronos-2 has identified. At $L=4096$ its projected gain is \KGainOne{} of the true one and its correlation with the true shape is \KRhoOne{} (1-DOF, medians over 30 instances; Fig.~\ref{fig:mechanism}c shows one instance). An ideal scalar correction of the predicted effect would lower the 1-DOF error only to \OracleGainOne.

\paragraph{Context excitation and output noise.} Across seven context excitation types at $L=1024$, Chronos-2's error ranks the excitations in nearly the same order as ARX's (Spearman \ExcRhoMin--\ExcRhoMax). Pseudo-random binary sequences (PRBS) and wide-band inputs work best (1-DOF: \EPrbsOne), and a single sine, which is not persistently exciting, fails for both (Chronos-2 \ESineOne, ARX \ESineOneArx). The model therefore faces the same identifiability limits as classical identification, without matching its accuracy. One dependence runs opposite to classical identification: noise in the context output \emph{improves} Chronos-2 (Fig.~\ref{fig:mechanism}d). On the 1-DOF class its error falls from \ENoiseZero{} to \ENoiseHigh{} as the noise grows from 0 to 30\% of the output standard deviation, while the ARX error rises from \ENoiseArxZero{} to \ENoiseArxHigh.

\section{Context dither at inference}\label{sec:dither}

We add Gaussian dither to the context output at inference (the future input is not altered) and average the forecasts over four draws, with common random numbers for the factual and alternative inputs of an instance. The dither level ($0.5\times$ the output standard deviation) was selected on 30 calibration instances per system that are not used anywhere else. Masking 25\% of the context output helped less (calibration median over systems \CalMaskQ{} against \CalDith{}, undisturbed \CalNone), and masking 75\% made it worse (\CalMaskH). On the test instances dither lowers the what-if error on all six classes, each with a paired 95\% interval below zero. On the 1-DOF class the error falls from \ECtwoOne{} to \EDithOne{} and the sensitivity rises to \SDithOne; on Wiener--Hammerstein the error falls from \ECtwoWh{} to \EDithWh. The improvement holds at $L=256$. It corrects the shape of the recovered impulse response (median correlation \KNoneRho{} to \KDithRho) as well as its gain (\KNoneGain{} to \KDithGain). A single draw gives \KoneShare\% of the averaged improvement, so the effect is not an ensemble effect. On the held-out hysteresis class dither lowers the error from \ECtwoBwOnly{} to \EDithBwOnly.

Three follow-up tests locate the effect. Dithering the input channel with white noise also helps (\NuwhiteE{} of 6 classes improved; 5 of 6 with an independent set of dither draws), but dither of the same power band-limited to the context input's bandwidth helps none and hurts \NubandWorse. The input-channel benefit therefore comes from widening the excitation band. After fine-tuning (Section~\ref{sec:adapt}), dither no longer helps (\NftdithE{} of 6 improved, \NftdithWorse{} worse). This is consistent with dither compensating a deficit of the pretrained model that fine-tuning removes. On the measured benchmarks it lowers the factual error of Chronos-2 on Wiener--Hammerstein (paired difference \MDdWh{} [\MDloWh, \MDhiWh]), Cascaded Tanks (\MDdCt{} [\MDloCt, \MDhiCt]; overlapping windows, Appendix~\ref{app:systems}) and Silverbox (\MDdSb{} [\MDloSb, \MDhiSb]), and it raises the shuffle gain. On our own base-excitation rig (Section~\ref{sec:adapt}) it raises the shuffle gain but worsens the factual error. Dither remains far from classical identification throughout.

\begin{table}[t]
\centering\small
\caption{Measured benchmarks: median factual NRMSE over windows ($L=1024$; 512 for Cascaded Tanks; $H=128$) and median shuffle gain (increase of NRMSE when the future input is taken from another window). NRMSE above 1 is worse than predicting the window mean. EMPS is excluded (periodic closed-loop record; Section~\ref{sec:protocol}). FT: training seed 1.}
\label{tab:bench}
\resizebox{\textwidth}{!}{\begin{tabular}{lcccccccccc}
\toprule
dataset & TimesFM-2.5 & TabPFN-TS & Chronos-2 & +dither & ICL-TF & FT & FT-wide & ARX & OE & NARX \\
\midrule
Wiener--Hammerstein (n=40) & 1.081 & 1.125 & 0.556 & 0.301 & 0.309 & 0.182 & 0.189 & 0.175 & 0.184 & 0.125 \\
\quad shuffle gain & +0.00 & +0.00 & +0.68 & +1.07 & +1.15 & +1.25 & +1.24 & +1.23 & +1.25 & +1.30 \\
Cascaded Tanks (n=18) & 1.215 & 1.417 & 1.197 & 0.774 & 1.553 & 0.428 & 0.421 & 0.541 & 0.495 & 0.521 \\
\quad shuffle gain & \ensuremath{-}0.01 & \ensuremath{-}0.01 & +0.23 & +0.66 & +0.16 & +1.08 & +1.14 & +1.06 & +1.45 & +0.85 \\
Silverbox (n=40) & 0.962 & 1.081 & 0.544 & 0.506 & 0.097 & 0.629 & 0.443 & 0.077 & 0.095 & 0.039 \\
\quad shuffle gain & +0.01 & +0.00 & +0.64 & +0.66 & +1.18 & +0.49 & +0.88 & +1.22 & +1.23 & +1.29 \\
\bottomrule
\end{tabular}}
\end{table}

\section{Fine-tuning on synthetic forced systems}\label{sec:adapt}

If attenuation were a capacity limit, more data of the right kind would not remove it. We fine-tuned Chronos-2 (FT) for \FTminutes{} minutes on a single Apple M4 Max GPU, using 3000 synthetic series from five of the six classes (Coulomb held out) with varied context excitation (narrow-band and sine excluded). The sensitivity rose to $\FTSrangeM$, and the error fell on every class (Table~\ref{tab:main}): on the 1-DOF class from \ECtwoOne{} to \FTmOne{} and on Wiener--Hammerstein from \ECtwoWh{} to \FTmWh{} (means over seeds). Across \NSeeds{} training seeds the system medians varied by at most \SeedRangeMax. A model fine-tuned only on the three linear classes also removed the attenuation on the nonlinear classes (Wiener--Hammerstein sensitivity \SFTlinWh), but its nonlinear errors were higher (\EFTlinWh). The attenuation is therefore mainly a deficit of the pretraining prior, and nonlinear accuracy additionally benefits from nonlinear classes in the prior. On the narrow-band context excitation, which was excluded from fine-tuning, the fine-tuned model improves on the base model in all three tested classes (lag+delay \HxNarrowLagCtwo{} to \HxNarrowLagFT, 1-DOF \HxNarrowOneCtwo{} to \HxNarrowOneFT, Duffing \HxNarrowDufCtwo{} to \HxNarrowDufFT). With a single-sine context, also excluded and tested on lag+delay, its error is \HxSineLagFT, where ARX reaches \HxSineLagArx{} and the base model \HxSineLagCtwo: the prior supplies what a context that is not persistently exciting cannot identify. On the narrow-band contexts ARX remains better (1-DOF \HxNarrowOneArx).

\begin{table}[t]
\centering\small
\caption{Fine-tuned Chronos-2 (seed-mean per instance) against the best structure-agnostic identifier on each class and context length: paired median difference of $\Eint$ over 30 instances with a 95\% bootstrap interval. Negative favours the fine-tuned model.}
\label{tab:ftcells}
\begin{tabular}{llcccll}
\toprule
system & $L$ & best agnostic & FT $-$ best & 95\% CI & per cell & Bonferroni \\
\midrule
lag+delay & 256 & NARX & +0.13 & [+0.12, +0.15] & worse & worse \\
1-DOF & 256 & ARX & +0.17 & [+0.15, +0.20] & worse & worse \\
2-DOF & 256 & OE & \ensuremath{-}0.03 & [\ensuremath{-}0.10, +0.09] & tie & tie \\
Duffing & 256 & ARX & \ensuremath{-}0.11 & [\ensuremath{-}0.17, \ensuremath{-}0.01] & better & better \\
Wiener--Ham. & 256 & ARX & \ensuremath{-}0.26 & [\ensuremath{-}0.35, \ensuremath{-}0.19] & better & better \\
Coulomb$^\dagger$ & 256 & ARX & \ensuremath{-}0.04 & [\ensuremath{-}0.08, \ensuremath{-}0.01] & better & tie \\
Bouc--Wen$^\dagger$ & 256 & OE & +0.01 & [\ensuremath{-}0.01, +0.07] & tie & tie \\
lag+delay & 1024 & NARX & +0.09 & [+0.08, +0.10] & worse & worse \\
1-DOF & 1024 & ARX & +0.15 & [+0.14, +0.18] & worse & worse \\
2-DOF & 1024 & N4SID & +0.13 & [\ensuremath{-}0.00, +0.18] & tie & tie \\
Duffing & 1024 & SE-NARX & +0.20 & [+0.03, +0.32] & worse & tie \\
Wiener--Ham. & 1024 & SE-NARX & \ensuremath{-}0.26 & [\ensuremath{-}0.30, \ensuremath{-}0.22] & better & better \\
Coulomb$^\dagger$ & 1024 & OE & \ensuremath{-}0.04 & [\ensuremath{-}0.07, \ensuremath{-}0.02] & better & better \\
Bouc--Wen$^\dagger$ & 1024 & NARX & \ensuremath{-}0.00 & [\ensuremath{-}0.05, +0.03] & tie & tie \\
\bottomrule
\end{tabular}
\end{table}

\paragraph{Comparison with structure-agnostic identification.} Table~\ref{tab:ftcells} compares the fine-tuned model with the best of six identifiers fitted to the same context, including SE-NARX. The intervals are per cell; the last column applies a Bonferroni correction over the 14 cells, under which \NLbetterB{} of the \NLbetter{} nonlinear advantages remain (Duffing at $L=256$ marginally: upper bound $-0.007$). At $L=256$ the simulation-error NARX was also run and did not improve on ARX (Duffing \SenSDuf, Wiener--Hammerstein \SenSWh, Coulomb \SenSCou), so the short-context comparisons on the nonlinear classes are against ARX or OE. On lag+delay and 1-DOF classical identification remains better (1-DOF: \FTmOne{} against \EBestOne, where ARX is the correct model class); on 2-DOF the comparison is a tie at both lengths. On the nonlinear classes the fine-tuned model is better in \NLbetter{} of \NLcells{} class--length cells, tied in \NLtie{} and worse in \NLworse. (Per-cell medians of the two models, as in Table~\ref{tab:main}, can differ from the paired median difference.) It is better on Wiener--Hammerstein and on the held-out Coulomb class at both lengths and on Duffing at $L=256$, tied on the held-out Bouc--Wen class, and worse only on Duffing at $L=1024$, where SE-NARX has the correct cubic structure for a cubic spring (Table~\ref{tab:main}: \BADuf{} against \FTmDuf). Table~\ref{tab:ftcells} therefore concerns identification without known physics; a grey-box model with the true equations is exact on Duffing, Wiener--Hammerstein and Coulomb (Table~\ref{tab:main}). The doubled input remains the hardest alternative after fine-tuning (Table~\ref{tab:alt}; Duffing \FTaltDufTwo), which is consistent with the arcsinh input compression that fine-tuning does not remove.

\paragraph{Training data and backbone.} Training the specialised in-context identification transformer~\citep{forgione2023incontext} on the same 3000 series (ICL-TF$^\ast$) brings it close to the fine-tuned Chronos-2 (Table~\ref{tab:main}; e.g.\ Wiener--Hammerstein \IoursWh{} against \FTmWh). The fine-tuned Chronos-2 is better with a paired interval below zero on \FTbeatsICLours{} of the 6 classes on which ICL-TF$^\ast$ was run, and it reads 1024 context samples where ICL-TF$^\ast$ reads 400. Most of the gain therefore comes from forced-system data in the prior. The general forecaster adds a modest advantage and keeps a single model for forecasting and what-if use.

\begin{figure}[t]
\centering
\includegraphics[width=\textwidth]{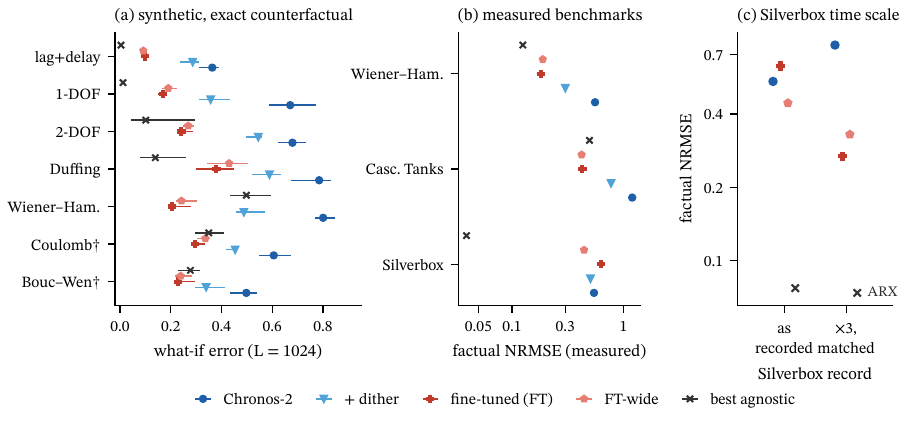}
\caption{(a) What-if error at $L=1024$: Chronos-2, context dither, fine-tuned (FT, three-seed mean), FT-wide (time-scale-randomised prior) and the best structure-agnostic identifier; medians, 95\% bootstrap intervals. (b) Factual NRMSE, measured benchmarks. (c) Silverbox as recorded and resampled threefold (context and horizon matched in physical time); classical reference ARX. In (b, c) FT is seed 1; log axes.}
\label{fig:remedies}
\end{figure}

\paragraph{Measured plants and prior coverage.} On the measured Wiener--Hammerstein benchmark the fine-tuned model reduces the error from \NCtwoWh{} to \NFTWh, which is still worse than the best classical model (\MBWh, \MBvWh; paired difference \MFTdWh{} [\MFTloWh, \MFThiWh]). On Cascaded Tanks its median error is lower (\NFTCt{} against \MBvCt{} for \MBCt), but the paired difference over windows is a tie (\MFTdCt{} [\MFTloCt, \MFThiCt]; overlapping windows, Appendix~\ref{app:systems}). On Silverbox it gets \emph{worse} than the base model (\NFTSb{} against \NCtwoSb), while ARX reaches \NArxSb. Silverbox's dominant resonance lies at about 0.11 cycles per sample (a period of about nine samples), below the modal periods of 10--60 samples in the fine-tuning classes (Appendix~\ref{app:systems}). This coverage explanation was tested three ways, on the same 40 windows. First, resampling the record threefold with context and horizon matched in physical time ($L = 3072$, $H = 384$) lowers the fine-tuned error to \RiFTm{} (FT-wide \RiFTwm), while the base model gets worse (\RiCtwom) and ARX is unchanged (\RiArxm). Second, the easier-task control of a short horizon on the original record ($H=43$) does not help the fine-tuned model relative to the base model (\RiFTs{} against \RiCtwos). Third, a second fine-tune with each training system time-scaled by a random factor between $1/3$ and $3$ (FT-wide) improves the original record to \NFTwSb, at a small cost on five of the six synthetic classes and on Bouc--Wen (Table~\ref{tab:main}). Coverage explains part of the Silverbox failure; most of the gap to ARX on the original time scale remains.

Our own base-excitation rig (a beam in stick-slip contact driven by a 10--100\,Hz sweep; base acceleration to beam velocity, 1024\,Hz after decimation; 40 windows, $L=1024$) is harder still. Each context window covers a narrow part of the sweep, so the contexts are weakly exciting; the sweep periods (10--100 samples) largely overlap the fine-tuning prior, so time-scale coverage is not the explanation here. Classical models reach an NRMSE of \SevOe{} (OE) and \SevArx{} (ARX), against \SevFT{} for the fine-tuned model (paired difference to OE \SevFTd{} [\SevFTlo, \SevFThi]), \SevCtwo{} for the base model and \SevTfm{} for TimesFM-2.5. Fine-tuning raises Chronos-2's shuffle gain from \SevGCtwo{} to \SevGFT, but on this plant classical identification is about four times more accurate.

\paragraph{Univariate forecasting after fine-tuning.} Fine-tuning specialises the model. On 200 hourly series of the M4 competition data~\citep{makridakis2020m4} (Appendix~\ref{app:systems}), a dataset used in neither fine-tuning corpus, the median mean absolute scaled error (MASE)~\citep{hyndman2006mase} of univariate forecasts rises from \MaseBase{} (base) to \MaseFT{} (fine-tuned; worse on \MaseWorsePct\% of series) and \MaseFTw{} (FT-wide). Mixing unrelated univariate series into the fine-tuning data did not prevent this (Appendix~\ref{app:replay}), so the adapted model is best used as a dedicated what-if model, consistent with catastrophic forgetting under full fine-tuning~\citep{kirkpatrick2017forgetting}.

\section{Related work}\label{sec:related}

\emph{Covariates in TSFMs.} Chronos-2~\citep{ansari2025chronos2}, TimesFM~\citep{das2024timesfm} and TabPFN-TS~\citep{hoo2025tabpfnts} accept future-known covariates and are benchmarked on factual accuracy. Berthelier et al.~\citep{berthelier2026covariates} probed static target--covariate relations. Our results add that, for dynamic relations, the interface decides what can be represented: a same-time regression cannot carry a delay or a resonance, however good the backbone.

\emph{Interventions and mechanisms.} Jander et al.~\citep{jander2026causal} intervene on the generator of a univariate series and find a persistence bias. Zhang and Gilpin~\citep{zhang2024zeroshot,zhang2025parroting} identify context parroting in autonomous chaotic systems; we find no parroting under exogenous input. For HVAC control, ThermoForce~\citep{wang2026thermoforce} separates a frozen TSFM free response from a learned forced-response operator after finding that covariate TSFMs fail on intervention effects. Our Chronos-2 results locate a similar failure in attenuation and show that it is largely repaired inside the foundation model on synthetic plants, though not on most measured ones.

\emph{In-context system identification.} Transformers meta-trained on synthetic system classes simulate new systems from context~\citep{forgione2023incontext,rufolo2024probabilistic}, can be adapted to new classes~\citep{piga2024adapting}, and dynamics foundation models can be pretrained purely on synthetic data~\citep{ziegler2024dynamics}. Our adaptation result is consistent with this line: a general forecaster needed forced-system data in its prior before its covariate channel behaved as an identifier on synthetic plants.

\section{Discussion}\label{sec:discussion}

Across the model--system cells of the synthetic study, factual NRMSE and what-if error are strongly correlated (Spearman \RhoFactWhatif). For the models that use the input the correlation also holds across the 30 instances of one model and system (Spearman \RkMin--\RkMax), because in a forced system driven by an unforeseen input the factual future is itself a response to input. Factual benchmarks with inputs are therefore informative for forced plants. Passive-dominated plants such as buildings differ: there, factual accuracy can hide invalid intervention responses~\citep{wang2026thermoforce}.

The two conditions identified here translate into a check before a covariate TSFM is used as the what-if model of a plant. Paired counterfactual inputs on synthetic plants of the plant's class and time scale measure the sensitivity and the response shape, and shuffled future inputs on the plant's own record show whether the model uses the input at all. A model that ignores the input has zero shuffle gain, and the two memoryless default interfaces have $-0.01$ to $+0.01$ (Table~\ref{tab:bench}); an attenuating prior has sensitivity below one. Exact counterfactuals exist only for the simulated classes: on a measured record the factual error and the shuffle gain show whether a model uses the input, not whether its predicted effect is correct. The synthetic classes cover linear, polynomial, saturating, friction and hysteresis nonlinearities at one sample-rate regime, without switching or strongly non-minimum-phase systems.

The Duffing and Wiener--Hammerstein test systems come from the training generator families (with disjoint seeds), so the synthetic advantage of the fine-tuned model on them is in-distribution. The held-out friction and hysteresis classes, the held-out excitations and the measured plants carry the out-of-distribution evidence. The main fine-tuned model was trained with \NSeeds{} seeds, the linear-only and wide-prior variants with one. When the equations of motion are known, a grey-box model, exact on the Duffing, Wiener--Hammerstein and Coulomb classes (Table~\ref{tab:main}), remains the better choice. The results refer to the package versions in Appendix~\ref{app:systems}, and the TimesFM-2.5 backbone was combined with a covariate interface other than its in-context regression only through engineered memory features (Appendix~\ref{app:e4}).

Output dither and white input dither improve Chronos-2, band-limited input dither does not, and dither stops helping after fine-tuning. These results are consistent with a mismatch between the context and the pretrained prior and, for the input channel, with the excitation bandwidth. They do not identify the internal mechanism.

\section{Conclusion}
For the forced engineering systems studied here, two properties of a covariate-aware foundation model were needed for useful what-if answers: a covariate interface that can represent dynamics, which a same-time regression cannot, and a prior that includes forced systems at the plant's time scale, which Chronos-2 lacks out of the box and a short synthetic fine-tune largely supplies on synthetic plants. The two properties were not sufficient: classical identification remained clearly better on the measured Wiener--Hammerstein benchmark, on Silverbox and on our rig, and the fine-tuned model lost part of its general forecasting skill. Factual benchmarks with inputs track what-if accuracy in forced systems, but only paired counterfactual inputs expose the attenuation.

\paragraph{Code and data.} The synthetic generator, seeds, model versions and training settings are given in Appendix~\ref{app:systems}, and the measured benchmarks are public~\citep{schoukens2026benchmarking}. Code, configurations and synthetic results are available from the corresponding author on request. The base-excitation rig records belong to a separate study and are not distributed.

\paragraph{Use of AI tools.} The author used AI coding and writing assistants as tools to implement and run the experiment and analysis code and to draft and edit the manuscript text. The author conceived the study, defined the research questions, designed the experiments and interpreted the results, and takes full responsibility for the content.

\bibliographystyle{unsrtnat}
\bibliography{references}

\begin{thebibliography}{23}
\providecommand{\natexlab}[1]{#1}
\providecommand{\url}[1]{\texttt{#1}}
\expandafter\ifx\csname urlstyle\endcsname\relax
  \providecommand{\doi}[1]{doi: #1}\else
  \providecommand{\doi}{doi: \begingroup \urlstyle{rm}\Url}\fi

\bibitem[Ansari et~al.(2025)Ansari, Shchur, K{\"u}ken, Auer, Han, Mercado,
  Rangapuram, Shen, Stella, Zhang, Goswami, Kapoor, Maddix, Guerron, Hu, Yin,
  Erickson, Desai, Wang, Rangwala, Karypis, Wang, and
  Bohlke-Schneider]{ansari2025chronos2}
Abdul~Fatir Ansari, Oleksandr Shchur, Jaris K{\"u}ken, Andreas Auer, Boran Han,
  Pedro Mercado, Syama~Sundar Rangapuram, Huibin Shen, Lorenzo Stella, Xiyuan
  Zhang, Mononito Goswami, Shubham Kapoor, Danielle~C. Maddix, Pablo Guerron,
  Tony Hu, Junming Yin, Nick Erickson, Prateek~Mutalik Desai, Hao Wang, Huzefa
  Rangwala, George Karypis, Yuyang Wang, and Michael Bohlke-Schneider.
\newblock {Chronos-2: From Univariate to Universal Forecasting}.
\newblock arXiv:2510.15821, 2025.

\bibitem[Das et~al.(2023)Das, Kong, Sen, and Zhou]{das2024timesfm}
Abhimanyu Das, Weihao Kong, Rajat Sen, and Yichen Zhou.
\newblock {A decoder-only foundation model for time-series forecasting}.
\newblock arXiv:2310.10688, 2023.

\bibitem[Hoo et~al.(2025)Hoo, M{\"u}ller, Salinas, and Hutter]{hoo2025tabpfnts}
Shi~Bin Hoo, Samuel M{\"u}ller, David Salinas, and Frank Hutter.
\newblock {From Tables to Time: Extending TabPFN-v2 to Time Series
  Forecasting}.
\newblock arXiv:2501.02945, 2025.

\bibitem[Berthelier et~al.(2026)Berthelier, Baranova, Pantea, Naour, Petralia,
  Nabil, and Palpanas]{berthelier2026covariates}
Gaspard Berthelier, Mariia Baranova, Andrei-Tiberiu Pantea, Etienne~Le Naour,
  Adrien Petralia, Tahar Nabil, and Themis Palpanas.
\newblock {Investigating simple target-covariate relationships for Chronos-2
  and TabPFN-TS}.
\newblock arXiv:2605.12200, 2026.

\bibitem[Jander et~al.(2026)Jander, van Heeswijk, and Mes]{jander2026causal}
Mathis Jander, Wouter van Heeswijk, and Martijn Mes.
\newblock {Causal Analysis for Time Series Foundation Models}.
\newblock arXiv:2608.24303, 2026.

\bibitem[Zhang and Gilpin(2024)]{zhang2024zeroshot}
Yuanzhao Zhang and William Gilpin.
\newblock {Zero-shot forecasting of chaotic systems}.
\newblock arXiv:2409.15771, 2024.

\bibitem[Zhang and Gilpin(2025)]{zhang2025parroting}
Yuanzhao Zhang and William Gilpin.
\newblock {Context parroting: A simple but tough-to-beat baseline for
  foundation models in scientific machine learning}.
\newblock arXiv:2505.11349, 2025.

\bibitem[Wang(2026)]{wang2026thermoforce}
Yifan Wang.
\newblock {ThermoForce: A Physics-Structured Interventional World Model for
  Building HVAC Control}.
\newblock arXiv:2607.03942, 2026.

\bibitem[Forgione et~al.(2023)Forgione, Pura, and Piga]{forgione2023incontext}
Marco Forgione, Filippo Pura, and Dario Piga.
\newblock {From system models to class models: An in-context learning
  paradigm}.
\newblock arXiv:2308.13380, 2023.

\bibitem[Rufolo et~al.(2024)Rufolo, Piga, Maroni, and
  Forgione]{rufolo2024probabilistic}
Matteo Rufolo, Dario Piga, Gabriele Maroni, and Marco Forgione.
\newblock {Enhanced Transformer architecture for in-context learning of
  dynamical systems}.
\newblock arXiv:2410.03291, 2024.

\bibitem[Piga et~al.(2023)Piga, Pura, and Forgione]{piga2024adapting}
Dario Piga, Filippo Pura, and Marco Forgione.
\newblock {On the adaptation of in-context learners for system identification}.
\newblock arXiv:2312.04083, 2023.

\bibitem[Ljung(1999)]{ljung1999system}
Lennart Ljung.
\newblock \emph{System Identification: Theory for the User}.
\newblock Prentice Hall, Upper Saddle River, NJ, 2 edition, 1999.

\bibitem[Van~Overschee and De~Moor(1994)]{vanoverschee1994n4sid}
Peter Van~Overschee and Bart De~Moor.
\newblock {N4SID}: Subspace algorithms for the identification of combined
  deterministic-stochastic systems.
\newblock \emph{Automatica}, 30\penalty0 (1):\penalty0 75--93, 1994.
\newblock \doi{10.1016/0005-1098(94)90230-5}.

\bibitem[Billings(2013)]{billings2013narmax}
Stephen~A. Billings.
\newblock \emph{Nonlinear System Identification: {NARMAX} Methods in the Time,
  Frequency, and Spatio-Temporal Domains}.
\newblock Wiley, 2013.
\newblock \doi{10.1002/9781118535561}.

\bibitem[Wen(1976)]{wen1976hysteretic}
Yi-Kwei Wen.
\newblock Method for random vibration of hysteretic systems.
\newblock \emph{Journal of the Engineering Mechanics Division}, 102\penalty0
  (2):\penalty0 249--263, 1976.
\newblock \doi{10.1061/JMCEA3.0002106}.

\bibitem[Schoukens et~al.(2026)Schoukens, Champneys, Beintema, and
  Rogers]{schoukens2026benchmarking}
Maarten Schoukens, Max Champneys, Gerben~Izaak Beintema, and Timothy~James
  Rogers.
\newblock Benchmarking for nonlinear system identification: Submission platform
  and baseline results.
\newblock \emph{Data-Centric Engineering}, 7, 2026.
\newblock ISSN 2632-6736.
\newblock \doi{10.1017/dce.2026.10067}.

\bibitem[Wigren and Schoukens(2013)]{wigren2013three}
Torbjorn Wigren and Johan Schoukens.
\newblock Three free data sets for development and benchmarking in nonlinear
  system identification.
\newblock In \emph{2013 European Control Conference (ECC)}, pages 2933--2938.
  IEEE, 2013.
\newblock \doi{10.23919/ecc.2013.6669201}.

\bibitem[Schoukens and No{\"e}l(2017)]{schoukens2017three}
Maarten Schoukens and Jean-Philippe No{\"e}l.
\newblock Three benchmarks addressing open challenges in nonlinear system
  identification.
\newblock \emph{IFAC-PapersOnLine}, 50\penalty0 (1):\penalty0 446--451, 2017.
\newblock \doi{10.1016/j.ifacol.2017.08.071}.

\bibitem[Hollmann et~al.(2025)Hollmann, M{\"u}ller, Purucker, Krishnakumar,
  K{\"o}rfer, Hoo, Schirrmeister, and Hutter]{hollmann2025tabpfn}
Noah Hollmann, Samuel M{\"u}ller, Lennart Purucker, Arjun Krishnakumar, Max
  K{\"o}rfer, Shi~Bin Hoo, Robin~Tibor Schirrmeister, and Frank Hutter.
\newblock Accurate predictions on small data with a tabular foundation model.
\newblock \emph{Nature}, 637\penalty0 (8045):\penalty0 319--326, January 2025.
\newblock ISSN 1476-4687.
\newblock \doi{10.1038/s41586-024-08328-6}.

\bibitem[Makridakis et~al.(2020)Makridakis, Spiliotis, and
  Assimakopoulos]{makridakis2020m4}
Spyros Makridakis, Evangelos Spiliotis, and Vassilios Assimakopoulos.
\newblock The {M4} competition: 100,000 time series and 61 forecasting methods.
\newblock \emph{International Journal of Forecasting}, 36\penalty0
  (1):\penalty0 54--74, 2020.
\newblock \doi{10.1016/j.ijforecast.2019.04.014}.

\bibitem[Hyndman and Koehler(2006)]{hyndman2006mase}
Rob~J. Hyndman and Anne~B. Koehler.
\newblock Another look at measures of forecast accuracy.
\newblock \emph{International Journal of Forecasting}, 22\penalty0
  (4):\penalty0 679--688, 2006.
\newblock \doi{10.1016/j.ijforecast.2006.03.001}.

\bibitem[Kirkpatrick et~al.(2017)Kirkpatrick, Pascanu, Rabinowitz, Veness,
  Desjardins, Rusu, Milan, Quan, Ramalho, Grabska-Barwinska, Hassabis, Clopath,
  Kumaran, and Hadsell]{kirkpatrick2017forgetting}
James Kirkpatrick, Razvan Pascanu, Neil Rabinowitz, Joel Veness, Guillaume
  Desjardins, Andrei~A. Rusu, Kieran Milan, John Quan, Tiago Ramalho, Agnieszka
  Grabska-Barwinska, Demis Hassabis, Claudia Clopath, Dharshan Kumaran, and
  Raia Hadsell.
\newblock Overcoming catastrophic forgetting in neural networks.
\newblock \emph{Proceedings of the National Academy of Sciences}, 114\penalty0
  (13):\penalty0 3521--3526, 2017.
\newblock \doi{10.1073/pnas.1611835114}.

\bibitem[Ziegler et~al.(2024)Ziegler, Posada-Moreno, Solowjow, and
  Trimpe]{ziegler2024dynamics}
Martin Ziegler, Andres~Felipe Posada-Moreno, Friedrich Solowjow, and Sebastian
  Trimpe.
\newblock {On Foundation Models for Dynamical Systems from Purely Synthetic
  Data}.
\newblock arXiv:2412.00395, 2024.

\end{thebibliography}

\appendix
\section{Systems, inputs and versions}\label{app:systems}

All synthetic systems are sampled at one sample per unit time with zero-order-hold input and integrated with fourth-order Runge--Kutta (8 sub-steps; 40 for the friction system). Initial transients are removed by a 600-sample burn-in. Output is standardised with the context mean and standard deviation, and the same transform is applied to the future. Parameters are sampled per instance ($\mathcal U$ = uniform):
\begin{itemize}
\item \textbf{lag+delay}: $y_{k+1} = a y_k + (1-a)K u_{k-d}$, $a = e^{-1/\tau}$, $\tau\sim\mathcal U(5,40)$, $d\in\{0,\dots,4\}$, $K\sim\mathcal U(0.5,2)$.
\item \textbf{1-DOF}: $\ddot x + 2\zeta\omega\dot x + \omega^2 x = u$, period $T\sim\mathcal U(16,48)$ samples, $\zeta\sim\mathcal U(0.03,0.2)$, $y=x$.
\item \textbf{2-DOF}: two parallel modes with periods $\mathcal U(30,60)$ and $\mathcal U(10,25)$, damping $\mathcal U(0.02,0.1)$, second-mode input gain $\mathcal U(0.5,1.5)$ and output weight $\pm\,\mathcal U(1,4)$.
\item \textbf{Duffing}: 1-DOF with period $T\sim\mathcal U(20,48)$ plus $\alpha x^3$, with $\alpha$ set so that at twice the linear response standard deviation the cubic force is $r\sim\mathcal U(0.5,2)$ times the linear spring force; $\zeta\sim\mathcal U(0.05,0.15)$.
\item \textbf{Wiener--Hammerstein}: resonant linear block ($T\sim\mathcal U(16,48)$, $\zeta\sim\mathcal U(0.1,0.3)$) $\to$ asymmetric saturation $f(v) = s\tanh(v/s)+\gamma v^2$ ($s\sim\mathcal U(0.5,1.5)$, $\gamma\sim\mathcal U(0,0.3)$) $\to$ first-order lag ($\tau\sim\mathcal U(3,15)$).
\item \textbf{Coulomb} (held out from all fine-tuning): oscillator with period $T\sim\mathcal U(20,48)$ and $\zeta\sim\mathcal U(0.02,0.1)$ plus smoothed Coulomb friction $F_c\tanh(\dot x/\epsilon)$, $F_c\sim\mathcal U(0.2,0.6)$, $\epsilon$ equal to 0.02 times the velocity standard deviation of the linear response, and an input dead zone with threshold $\delta\sim\mathcal U(0,0.3)$, i.e.\ the input $\mathrm{sign}(u)\max(|u|-\delta, 0)$.
\item \textbf{Bouc--Wen} (test only): $\ddot x + 2\zeta\omega\dot x + \omega^2(a x + (1-a) z) = u$, $\dot z = \dot x - \beta|\dot x| z - \gamma \dot x |z|$, period $T\sim\mathcal U(20,48)$, $\zeta\sim\mathcal U(0.02,0.1)$, $a\sim\mathcal U(0.1,0.4)$, $\beta\sim\mathcal U(2,6)/x_s$, $\gamma\sim\mathcal U(-1,1)/x_s$ with $x_s$ the linear response standard deviation.
\end{itemize}
The context input is zero-mean Gaussian noise passed through a fourth-order Butterworth low-pass filter with cutoff $\min(0.4,\max(0.08, 3f_{\mathrm{res}}))$ cycles/sample, scaled to unit variance. In the excitation study the context input was instead a narrow-band signal (cutoff $0.5 f_{\mathrm{res}}$), a wide-band signal (0.45), a multisine (30 random tones), a PRBS, random steps, or a single sine at $0.7 f_{\mathrm{res}}$. Seeds are derived with \texttt{numpy.random.SeedSequence} from integer tuples (system code, $L$, $H$, instance index). Evaluation instances (0--29), dither calibration instances (100--129) and fine-tuning series (a separate root seed) do not overlap.

\paragraph{Model versions.} chronos-forecasting 2.3.2 (\texttt{amazon/chronos-2}), timesfm 3.0.2 (\texttt{google/timesfm-2.5-200m-pytorch}), tabpfn-time-series 1.3.0 with \texttt{Prior-Labs/TabPFN-v2-reg}, sysid-transformers (\texttt{github.com/forgi86/sysid-transformers}, commit \texttt{2420b67}, checkpoints v0.3), nonlinear-benchmarks 1.0.1. Inference and fine-tuning ran on one Apple M4 Max (Metal backend).

\paragraph{Context-length, excitation and noise study.} The context-length, excitation and noise studies of Section~\ref{sec:diagnosis} (Fig.~\ref{fig:mechanism}) use 30 further instances per class drawn from a separate root seed, context lengths from 128 to 8192 samples and five alternative inputs: zero input, a step added, a resonant sine, a fresh input realisation and a pulse of amplitude 3 added at horizon step 8. The instance score is the median $\Eint$ over the five. The recovered impulse response $\hat h$ is the predicted effect of the pulse from step 8 on, divided by 3; it is compared with the unit impulse response $h$ of the system linearised at zero through the projected gain $\langle\hat h, h\rangle/\langle h, h\rangle$ and the Pearson correlation of $\hat h$ and $h$. The output-noise sweep uses $L=1024$ and noise levels 0, 0.03, 0.1 and 0.3 of the output standard deviation.

\paragraph{Context masking.} In the masking variants of the dither calibration, a random fraction (25\% or 75\%) of the context output samples is set to missing; Chronos-2 handles missing values natively.

\paragraph{Fine-tuning.} Full fine-tuning of Chronos-2 with learning rate $10^{-5}$, 2000 steps and batch size 64, context 1024 and horizon 128, on 600 series of length 2176 per class (lag+delay, 1-DOF, 2-DOF, Duffing, Wiener--Hammerstein). Context excitations were drawn from the band-limited, wide-band, multisine, PRBS and step signals; narrow-band and sine contexts were excluded so that they could be tested as held-out excitations. Output noise was drawn from $\{0, 0.03, 0.1\}$. The linear-only variant used 1000 series of each linear class. The wide-prior variant additionally time-scales each training system by $c\sim\log\mathcal U(1/3,3)$.

\paragraph{Retrained in-context identifier (ICL-TF$^\ast$).} The architecture and checkpoint format of ICL-TF are kept. Training uses random contiguous segments of 400 context and 100 query samples cut from the 3000 FT training series, with $y$ standardised on the context part, the simulation mean squared error on the query, AdamW (betas 0.9 and 0.95) with learning rate $10^{-4}$, 100 warm-up steps and cosine decay to $10^{-5}$, gradient-norm clipping at 1.0, batch size 32 and 1600 steps. The checkpoint with the lowest error on 500 fixed segments from 100 further series of the same generator (disjoint seeds), evaluated every 100 steps, is used.

\paragraph{Grey-box fitting.} For Duffing, Wiener--Hammerstein and Coulomb the grey-box model has the true equations with unknown physical parameters, initial state, output scale and offset. They are fitted by nonlinear least squares on the free-run simulation error over the context, from six random starts drawn from the class prior (150 function evaluations each), and the best fit is kept.

\paragraph{Univariate forecasting test.} The 200 M4 hourly series are drawn at random with a fixed seed. Each series is forecast 48 samples ahead from its full preceding history, and MASE is scaled by the in-sample seasonal naive error with a lag of 24 samples.

\paragraph{Alternative inputs.} The resonant alternative is $\sqrt2\sin(2\pi f_{\mathrm{res}} t)$ with $f_{\mathrm{res}}$ the dominant frequency: the (higher) modal frequency for the oscillators and the corner frequency $1/(2\pi\tau)$ for lag+delay. The step alternative adds $+1$ from $t = H/4$. The fresh alternative is an independent realisation of the context input process.

\paragraph{Memorylessness statistic.} For each instance and alternative, the per-sample effect $\Delta\hat y_t$ is regressed on the same-sample input change $\Delta u_t$ with an intercept over the $H$ horizon samples. $R^2$ is the median over the six alternatives and then over instances. The same statistic computed for the true effect $\Delta y_t$ is reported as the reference.

\paragraph{Classical baselines.} ARX: $n_a, n_b \in\{1,2,3,4,6,8\}$, delay $n_k\in\{1,\dots,5\}$, ridge $10^{-6}$. OE: initialised from the selected ARX and refined by Levenberg--Marquardt on simulation error. N4SID: order $\in\{1,2,3,4,5,6,8\}$, 15 block rows; the initial state at the forecast origin is estimated by least squares over the last 200 context samples. Polynomial NARX: $n$ lags of $y$ and $u$ with $n\in\{2,3,4,6\}$, degree $\le 3$, ridge $10^{-4}$, fitted by one-step least squares and selected by free-run simulation error on the last quarter of the context. Simulation-error NARX refines the selected polynomial NARX by least squares on free-run simulation error over the whole context (10 function evaluations; $L = 1024$ for all nonlinear classes, and $L = 256$ for Duffing, Wiener--Hammerstein, Coulomb and Bouc--Wen). MLP-NARX: two hidden layers of 64 tanh units, lags $n\in\{4,8\}$, L2 penalty $\in\{10^{-4},10^{-2}\}$, three initialisations, one-step training with early stopping and selection by free-run simulation error.

\paragraph{Intervals.} All intervals are 95\% percentile bootstrap intervals over instances (or windows) with fixed seeds. The paired differences of Table~\ref{tab:ftcells} and of the public measured benchmarks use 20{,}000 resamples; the other intervals use 2000.

\paragraph{Measured windows.} Windows start at evenly spaced positions of the training record: 40 non-overlapping windows for Silverbox and Wiener--Hammerstein; 9 windows each from the training and test records of Cascaded Tanks ($L = 512$), which overlap within each 1024-sample record, so their intervals are optimistic. The shuffled input of window $w$ is the future input of window $w + n/2$ (mod $n$).

\section{Additional results}

\begin{table}[h]
\centering\small
\caption{As Table~\ref{tab:main} at $L = 256$.}
\label{tab:main256}
\resizebox{\textwidth}{!}{\begin{tabular}{lcccccccccc}
\toprule
system & TimesFM-2.5 & TabPFN-TS & Chronos-2 & +dither & ICL-TF & ICL-TF$^\ast$ & FT & FT-wide & best agnostic & known struct. \\
\midrule
lag+delay & 1.00 & 1.01 & 0.86 & 0.39 & 0.42 & 0.19 & 0.14$\pm$0.01 & 0.14 & 0.01 (NARX) & -- \\
1-DOF & 1.01 & 1.00 & 0.84 & 0.45 & 0.25 & 0.27 & 0.21$\pm$0.01 & 0.24 & 0.03 (ARX) & -- \\
2-DOF & 1.00 & 1.00 & 0.89 & 0.62 & 0.51 & 0.44 & 0.35$\pm$0.02 & 0.37 & 0.37 (OE) & -- \\
Duffing & 1.00 & 1.01 & 0.89 & 0.63 & 0.75 & 0.52 & 0.50$\pm$0.01 & 0.54 & 0.62 (ARX) & 0.01 \\
Wiener--Ham. & 1.01 & 1.00 & 0.93 & 0.56 & 0.79 & 0.31 & 0.27$\pm$0.01 & 0.29 & 0.54 (ARX) & 0.00 \\
Coulomb$^\dagger$ & 1.00 & 1.00 & 0.77 & 0.53 & 0.61 & 0.39 & 0.38$\pm$0.01 & 0.42 & 0.48 (ARX) & 0.01 \\
Bouc--Wen$^\dagger$ & -- & -- & 0.87 & 0.48 & -- & -- & 0.35$\pm$0.00 & 0.33 & 0.34 (OE) & -- \\
\bottomrule
\end{tabular}}
\end{table}

\begin{table}[h]
\centering\small
\caption{What-if error / sensitivity per alternative input ($L=1024$, medians over 30 instances).}
\label{tab:alt}
\resizebox{\textwidth}{!}{\begin{tabular}{llcccccc}
\toprule
model & system & $\times0.5$ & $\times2$ & zero & step & resonant & fresh \\
\midrule
Chronos-2 & lag+delay & 0.37/0.85 & 0.52/0.70 & 0.32/0.87 & 0.37/0.66 & 0.38/0.70 & 0.30/0.86 \\
Chronos-2 & 1-DOF & 0.68/0.57 & 0.78/0.35 & 0.66/0.65 & 0.66/0.48 & 0.66/0.39 & 0.61/0.54 \\
Chronos-2 & 2-DOF & 0.71/0.67 & 0.84/0.41 & 0.68/0.65 & 0.53/0.66 & 0.67/0.43 & 0.61/0.58 \\
Chronos-2 & Duffing & 0.82/0.49 & 0.96/0.33 & 0.75/0.55 & 0.69/0.46 & 0.71/0.53 & 0.67/0.50 \\
Chronos-2 & Wiener--Ham. & 0.82/0.42 & 0.95/0.20 & 0.75/0.48 & 0.83/0.24 & 0.76/0.35 & 0.76/0.35 \\
Chronos-2 & Coulomb$^\dagger$ & 0.64/0.57 & 0.80/0.33 & 0.50/0.75 & 0.48/0.68 & 0.64/0.41 & 0.53/0.70 \\
Chronos-2 & Bouc--Wen$^\dagger$ & 0.55/0.71 & 0.83/0.38 & 0.47/0.78 & 0.40/0.66 & 0.48/0.74 & 0.43/0.71 \\
FT & lag+delay & 0.11/0.96 & 0.16/0.98 & 0.09/0.98 & 0.07/0.96 & 0.09/0.94 & 0.06/0.98 \\
FT & 1-DOF & 0.19/0.91 & 0.28/0.87 & 0.14/0.94 & 0.13/0.93 & 0.20/0.86 & 0.12/0.94 \\
FT & 2-DOF & 0.25/0.98 & 0.33/0.93 & 0.22/0.98 & 0.17/0.97 & 0.33/0.81 & 0.20/0.95 \\
FT & Duffing & 0.38/0.89 & 0.70/0.72 & 0.26/0.94 & 0.39/0.88 & 0.35/0.94 & 0.24/0.96 \\
FT & Wiener--Ham. & 0.25/0.91 & 0.45/0.86 & 0.18/0.96 & 0.15/0.96 & 0.26/0.93 & 0.19/0.97 \\
FT & Coulomb$^\dagger$ & 0.40/0.69 & 0.47/0.64 & 0.26/0.91 & 0.20/0.92 & 0.33/0.75 & 0.24/0.93 \\
FT & Bouc--Wen$^\dagger$ & 0.30/0.89 & 0.38/0.86 & 0.25/0.94 & 0.14/0.96 & 0.27/1.01 & 0.23/0.95 \\
\bottomrule
\end{tabular}}
\end{table}

\paragraph{Memory covariates.}\label{app:e4} Median $\Eint$ at $L=1024$ with 32 lagged inputs as extra covariates. TimesFM-2.5: \ETfmLagMin--\ETfmLagMax{} across classes. Chronos-2: 1-DOF \ECtwoLagOne{} (base \ECtwoOne). A causal first-order filter bank with 14 channels (low-pass outputs with time constants 2, 4, 8, 16, 32, 64 and 128 samples, and seven band-pass differences between successive outputs, starting with the input minus the first low-pass output) gave TimesFM-2.5 \ETfmFbMin--\ETfmFbMax{} and Chronos-2 \ECtwoFbMin--\ECtwoFbMax. No feature set made a model better than the best classical identifier on any class (paired 95\% interval below zero).

\paragraph{Replay against forgetting.}\label{app:replay} \ReplayText

\end{document}